# Enhancing Annotated Bibliography Generation

# with LLM Ensembles

*Sergio Bermejo*

Departament d'Enginyeria Electrònica,

Universitat Politècnica de Catalunya (UPC),

Jordi Girona 1-3 (C4 building), 08034 BARCELONA, SPAIN

PHONE: +34 4016758, E-MAIL: sergio.bermejo@upc.edu

**Abstract.** This work proposes a novel approach to enhancing annotated bibliography generation through Large Language Model (LLM) ensembles. In particular, multiple LLMs in different roles—controllable text generation, evaluation, and summarization—are introduced and validated using a systematic methodology to enhance model performance in scholarly tasks. Output diversity among the ensemble that generates text is obtained using different LLM parameters, followed by an LLM acting as a judge to assess relevance, accuracy, and coherence. Responses selected by several combining strategies are then merged and refined through summarization and redundancy removal techniques. The preliminary experimental validation demonstrates that the combined outputs from the LLM ensemble improve coherence and relevance compared to individual responses, leading to a 38% improvement in annotation quality and a 51% reduction in content redundancy, thus highlighting the potential for automating complex scholarly tasks while maintaining high-quality standards.

**Keywords**: Large Language Models, Ensemble Learning, Automated Bibliography Generation, Scholarly Writing, Artificial Intelligence, Knowledge Synthesis.

# 1. Introduction

Annotated bibliographies are research tools that summarize and critically assess the relevance, accuracy, and quality of sources [1, 2]. This critical evaluation—ideally concise but accurate—differentiates annotations from abstracts and thus makes them valuable for researchers [3]. The advent of large language models (LLMs) [4, 5], which have already transformed many sectors including education and scholarly research [6, 7], offers the potential to automate such complex tasks that require natural language understanding and human expertise [8]. Although previous research has highlighted the potential of LLMs in generating textual content [9], their application in structured academic outputs like annotated bibliographies has been overlooked.

Some benefits of LLMs [10] for generating annotated bibliographies are the acceleration in annotation and summarization processes compared to manual approaches since they can analyze large amounts of text quickly, and improved consistency and objectivity compared to human experts because LLMs can use predefined criteria and guidelines consistently across all annotations, thus ensuring the reliability and comparison of annotations.

However, individual LLMs often face limitations in terms of accuracy and bias when generating content due to the generated output's dependence on training data biases [11], the architecture and parameters of LLMs [12,13], and the context and message provided in the prompts [14]. Ensemble methods for LLMs [15] have emerged as a promising approach to reducing the effect of such dependencies, and thus improving the accuracy and reliability of content generation.

This work explores the enhancement of annotated bibliography generation by utilizing LLM ensembles. The strengths of multiple LLMs in different roles—controllable text generation [16], evaluation [17], and summarization [18][19]—are combined to improve the quality and reliability of generated annotations. Also, various ensemble techniques [20]—including voting and averaging—have been investigated and evaluated for their effectiveness in generating comprehensive and insightful annotations.

The paper's organization is as follows: Section 2 presents the background and methodology, Section 3 details the experimental setup, results, and analysis of the findings, and Section 4 concludes with recommendations for future research directions.

## 2. Annotated Bibliographies using LLM ensembles

Annotated bibliographies are citations accompanied by concise summaries and critical evaluations, providing researchers with an overview of relevant literature to aid in source selection and literature review [21]. The annotations include a descriptive summary containing key findings of sources and an evaluation that helps to discern their relevance [22]. LLMs, such as GPT-4 [23], which have demonstrated remarkable capabilities in processing, understanding, and generating large amounts of text, are excellent candidates for automating the generation of annotated bibliographies. However, the overall accuracy and reliability of LLM content generation can be significantly enhanced by combining the outputs of multiple LLMs that form an ensemble [15].

The inherent challenges of automated systems for annotated bibliography tasks are particularly demanding:

● *Complex Evaluation Criteria*: Annotated bibliography generation requires evaluation beyond simple text quality. An automated system needs to assess factors such as the relevance of the references to the topic, the accuracy and completeness of the annotations, and the overall coherence and structure of the bibliography.

● *Subjectivity and Interpretation*: Annotations often involve subjective judgments and interpretations of the source material. This introduces a degree of subjectivity into the evaluation process, making it challenging to define absolute quality standards.

● *Domain Expertise*: Assessing the relevance and accuracy of references in an annotated bibliography ideally requires domain expertise on the topic. While an automated system can have broad knowledge, it may not always have the depth of expertise needed for evaluation in specific domains.

In this work, a new architecture that uses multiple LLMs in different roles within a three-tier LLM chain ensemble [24] is proposed to address the above challenges:

*Level 1. Controllable text generation*: To maximize diversity in outputs [25,] Multiple LLMs with different hyperparameter settings (temperature, top_k, and top_p) are used to generate text responses, i.e. annotated bibliography entries; this exploration aims to obtain diverse responses with different characteristics. A particular optimization of the LLM hyperparameters is additionally done to obtain enhanced outputs for the annotated bibliography generation task.

*Level 2. Evaluation*: An LLM, acting as a judge [26], is then used to assess the relevance, accuracy, and coherence of the generated annotated bibliographies since the LLM-as-a-judge approach can achieve greater objectivity than that obtained using traditional metrics [27]. Hence, the generated responses are presented to an LLM acting as a judge along with the original prompt to evaluate the quality of each response and assign numerical ratings based on criteria relevant to annotated bibliography generation (relevance, accuracy, coherence, etc.).

*Level 3. Summarization*: Responses selected by several combining strategies are finally merged and refined through summarization and redundancy removal techniques through a third LLM. In particular,

- Rating analysis and selection is first done by extracting the ratings from LLM-as-a-judge's response and calculating, e.g., average and majority votes for each parameter configuration; two straightforward approaches are used to select the best responses: the top temperature approach, which selects responses generated with the temperature that received the highest average rating, and the top M responses approach, which chooses the top M responses based on their ratings [25][28].

- Response combination and refinement are then done by summarizing the answers chosen in the two approaches using a third LLM, in which redundancy is removed from the summaries using sentence similarity techniques; finally, the remaining information is combined to produce a final higher-quality annotated bibliography output.

The following variables may influence the outcomes of this architecture:

● *Model Parameters*: Variations in LLM hyperparameters such as temperature, top_k, and top_p directly affect the diversity and quality of generated outputs.

● *Quality Assessment Criteria*: The criteria used by the LLM acting as a judge are essential for evaluating the effectiveness of generated annotations.

● *Ensemble method*: The ensemble strategies employed (e.g., selection of top responses based on ratings) will affect the final quality of the annotated bibliography.

● *Domain Expertise*: The relevance of the annotated bibliographies generated will depend on the LLM's knowledge of the topic, impacting both the quality of the annotations provided and the accuracy with which those annotations are subsequently evaluated.

## 3. Results and discussion

In several preliminary experiments using Gemini 1.5 flash and Gemini 1.5 pro, LLM ensembles outperformed individual models for high-quality annotated bibliographies (see Table 1) since the Top M Responses and Top Temperature methods far outperform the baseline individual LLM and the Mean Individual LLM results in terms of both average sentence length and readability score.

Top M produced the most readable output, with an average readability score of 31.41, achieving a 38% improvement over the baseline readability score of 22.71 and a 23% improvement over the mean individual readability score of 25.01. The Top Temperature method also showed satisfactory results, with a 17% improvement over the baseline and a 6% improvement over the mean individual score. Hence, Top M produced more readable output than Top Temperature presumably because it selected responses based on their individual ratings, which included factors such as coherence and organization, while Top Temperature selected responses based solely on the temperature setting, which may not have captured all aspects of readability. However, Top Temperature outperforms Top M in conciseness, achieving a 51% (vs. 44%) reduction compared to the baseline and a 45% (vs. 35%) reduction compared to the mean individual result.

These improvements in readability and conciseness demonstrate the effectiveness of using LLM ensembles for annotated bibliography generation. By combining multiple LLM models and employing effective selection and refinement techniques, the ensemble methods can produce higher-quality and more informative annotations than individual LLMs.

INSERT TABLE 1

The experimental results demonstrate the effectiveness of using LLM ensembles for generating annotated bibliographies. By using the proposed three-tier LLM architecture, the ensemble methods produced higher quality and informative annotations than individual LLMs. Additionally, the analysis using various metrics provided insights into the strengths and weaknesses of the two approaches—Top M Responses and Top Temperature—for selecting and combining responses.

The variation in the LLM hyperparameters allows the generation of different annotations, followed by their effective evaluation done by LLM-as-a-judge, which assigns numerical ratings that reflect their quality, relevance, coherence, and factual accuracy. This LLM-as-a-judge approach enabled the identification of optimal parameter configurations that consistently produced better results, demonstrating the potential of LLMs for evaluating and guiding the optimization of other LLMs. Besides, the selection of high-quality responses based on ratings, followed by summarization and redundancy removal, produced a combined output superior to individual responses. This process highlights the importance of response combination and refinement in enhancing the quality of LLM-generated content.

Overall, this work shows the potential of LLM ensembles in automating complex scholarly tasks like annotated bibliography generation. Multiple LLMs followed by an effective evaluation enable the generation of high-quality annotations that can significantly enhance research productivity. However, further research is needed to address the different limitations present in the current work.

## 4. Conclusions

The proposed three-tier LLM architecture for automated annotated bibliography generation has shown promising results, increasing efficiency and higher-quality outputs in scholarly tasks. The results highlight the potential of LLM ensembles to automate complex scholarly tasks, indicating that the enhancement in the quality of LLM-generated content is strongly influenced by the methods of response combination and refinement.

Future work will address the limitations of the current LLM ensemble related to the evaluation criteria used by the LLM judge, the selection strategies for combining responses, and potential biases within the LLMs. Additionally, the current experimental setup will be refined, and more sophisticated ensemble techniques will be introduced.

| Output | Avg. Sentence Length | Readability Score |
|---|---|---|
| Baseline (Individual) | 39.00 | 22.71 |
| Mean Individual (vs. Baseline) | 34.80 | 25.01 |
| Top M Responses | 22.80 | 31.41 |
| Top Temperature | 19.11 | 26.71 |

Table 1: Comparison of LLM ensemble methods and baseline in terms of average sentence length and readability score.